\pdfoutput=1
\documentclass{article} 
\usepackage{nips15submit_e,times}
\usepackage{hyperref}
\usepackage{multirow}
\usepackage{multicol}
\usepackage{subfigure}
\usepackage{graphicx}
\usepackage[numbers]{natbib}
\usepackage{url}
\usepackage{amsmath, amssymb}

\newcommand{\field}[1]{\mathbb{#1}}
\newcommand{\R}{\field{R}} 

\newcommand{\vct}[1]{\boldsymbol{#1}} 
\newcommand{\mat}[1]{\boldsymbol{#1}} 
\newcommand{\T}{^{\textrm T}} 

\title{Distilling Knowledge from Deep Networks with Applications to Healthcare Domain}

\author{
Zhengping Che*, Sanjay Purushotham*,  Robinder Khemani**, Yan Liu*\\
*Department of Computer Science,  University of Southern California\\
**Children's Hospital Los Angeles\\
*\texttt{\{zche, spurusho, yanliu.cs\}@usc.edu}, **\texttt{RKhemani@chla.usc.edu}
}

\nipsfinalcopy 

\begin{document}

\maketitle

\begin{abstract}
Exponential growth in Electronic Healthcare Records (EHR) has resulted in new opportunities and urgent needs for discovery of meaningful data-driven representations and patterns of diseases in \textit{Computational Phenotyping} research. Deep Learning models have shown superior performance for robust prediction in computational phenotyping tasks, but suffer from the issue of model interpretability which is crucial for clinicians involved in decision-making. In this paper, we introduce a novel knowledge-distillation approach called \textbf{Interpretable Mimic Learning}, to learn interpretable phenotype features for making robust prediction while mimicking the performance of deep learning models.
Our framework uses Gradient Boosting Trees to learn interpretable features from deep learning models such as Stacked Denoising Autoencoder and Long Short-Term Memory. 
Exhaustive experiments on a real-world clinical time-series dataset show that our method obtains similar or better performance than the deep learning models, and it provides interpretable phenotypes for clinical decision making.

\end{abstract}

\section{Introduction}
\label{sec:introduction}


With the exponential surge in the amount of electronic health records (EHR) data, there come both the opportunities and the urgent needs for discovering meaningful data-driven characteristics and patterns of diseases, which is known as \textit{phenotype discovery}. Clinicians are collaborating with machine learning researchers to tackle many computational phenotyping problems to improve the state of healthcare services and this is paving the way for \textit{Personalized Healthcare}~\cite{chawla2013bringing}.
Robust prediction is critical in healthcare research since it is directly related to saving patient lives. Recent works~\cite{che2015deep, lasko2013computational}  have used deep learning models to achieve state-of-the-art performance on computational phenotype prediction problems. However, deep learning models are less interpretable while clinicians mainly rely on interpretable models to make informed clinical decisions. Thus, the fundamental question is how we can develop new data-driven machine learning techniques which can achieve state-of-the-art performance as deep learning models and also discover interpretable features (phenotypes).



Deep learning models are revolutionizing many fields such as computer vision~\cite{krizhevsky2012imagenet,karpathy2014deep, xu2015show}, and speech and language processing~\cite{mikolov2013distributed, zeiler2013rectified}, and have achieved the status as the go-to state-of-the-art techniques for many machine learning tasks. 
With the flexibility and power of all kinds of neural networks, some deep network variants are also potentially suitable for healthcare tasks~\cite{che2015deep, lipton2015learning}.
Autoencoder~\cite{vincent2008extracting} is used to capture structures in data and aims to reconstruct the input. It has been successfully used for feature extraction or as a pre-training step for a neural network~\cite{rifai2011contractive, hinton2012deep}.
Long Short-Term Memory (LSTM) architecture~\cite{hochreiter1997long} has been widely used for sequential data and tasks recently~\cite{sutskever2014sequence, bahdanau2014neural}. It reads the input sequence one time step at a time and provides fixed-length or step-by-step representations, while keeping the long range temporal dependencies.
While deep models perform superior to many approaches when the data is abundant, their performance can drop significantly when the data is noisy and sparse or when the model is not properly initialized~\cite{erhan2010does}. Also, the features learned using deep models are generally not interpretable.

There is limited work on interpreting the features learned by deep learning outside of computer vision~\cite{erhan2009visualizing}. Recent research has begun to provide a more rigorous understanding of the representations learned by deep architectures. It has been shown that the semantics encoded by hidden unit activations in one layer are preserved when projected onto random bases, instead of the next layer's bases~\cite{szegedy2013intriguing}. This implies that the practice of interpreting individual units can be misleading and that the behavior of deep models may be more complex than previously believed.  In healthcare, model interpretability is not only important but also \textit{necessary}, since the primary care providers, physicians and clinical experts alike depend on the new healthcare technologies to help them in monitoring and decision-making for patient care. A good interpretable model is shown to result in faster adoptability among the clinical staff and results in better quality of patient care~\cite{peleg2003comparing, kerr2012further}. Therefore we need to identify novel solutions which can provide interpretable models and achieve similar prediction performance as deep models in healthcare domain.


In order to capture the performance of deep learning models using other models, a knowledge distillation process such as mimic learning~\cite{ba2014deep} or dark knowledge~\cite{hinton2015distilling, korattikara2015bayesian}  is useful. In these works, it was noted that once a deep learning model is trained on a large-scale dataset, we can use a smaller model to distill the knowledge by training it on the ``soft target'', i.e., the class probability produced by the former model. This means that simple models can possibly match (mimic) the prediction performance of deep models. Thus, choosing the right simple model will help to extract informative physiologic patterns which in-turn helps to discover meaningful interpretable features.

Building upon the recent breakthrough in mimic learning, in this paper, we introduce our knowledge-distillation approach called \textbf{Interpretable Mimic Learning}, to learn interpretable features for making robust prediction while mimicking the performance of deep learning models. Unlike the standard mimic learning~\cite{ba2014deep}, our interpretable mimic learning framework uses Gradient Boosting Trees (GBT) to learn interpretable features from deep learning models.  We use GBT as our mimicking model since they not only provide interpretable decision rules and tree structures, but also
successfully maintain the performance of original complex models such as deep networks. Our main contributions in this paper include:
\begin{itemize}
	\item We propose a novel knowledge distillation methodology called \textit{Interpretable Mimic Learning} where we mimic the performance of state-of-the-art deep learning models using well-known Gradient Boosting Trees (GBT).
	\item We conduct extensive experiments on several deep learning architectures including Stacked denoising autoencoders (SDA) and Long Short Term Memory (LSTM) to show that our Interpretable Mimic Learning models can achieve state-of-the-art performance on multiple deep learning models.
	\item We discuss the interpretable features and decision rules learned by our Interpretable Mimic Learning models, which is validated by the expert clinicians.
We also conduct experiments to investigate, for different deep networks, whether using neural network extracted features rather than soft labels improves mimicking performance.
\end{itemize}

The remainder of this paper is arranged as follows: In Section~\ref{sec:relatedwork}, we provide an overview of the related work; In Section~\ref{sec:methods}, we describe our proposed Interpretable Mimic Learning framework and discuss the related deep learning models; An evaluation on empirical results and interpretable features is presented in the Section~\ref{sec:experiments}; We conclude with discussion, summary and future work in the Section~\ref{sec:discussions}.

\section{Related Work}
\label{sec:relatedwork}


In this section, we first provide an overview of the state-of-the-art deep learning approaches used in the healthcare domain, and then we discuss the recent advances in \textit{Mimic learning} approaches.
Recently, there is a growing interest in applying deep learning techniques to computational phenotyping~\cite{oellrich2015digital} due to the increasing availability of the Electronic Healthcare Records (EHR) and the need for Personalized Healthcare~\cite{belle2013biomedical, hamburg2010path}. One of the first applications of modern deep learning to clinical time series was described in~\cite{lasko2013computational}, where the authors use autoencoders to learn features from longitudinal clinical measurement time series and show interpretable features which are useful for classifying and clustering different types of patients. In our previous work~\cite{che2015deep}, we proposed a novel scalable deep learning framework which models the prior-knowledge from medical ontologies to learn interpretable and clinically relevant features for patient diagnosis in Intensive Care Units (ICU).  A recent study~\cite{dabek2015neural} showed that a neural network model can improve the prediction of the likelihood of several psychological conditions such as anxiety, behavioral disorders, depression, and post-traumatic stress disorder. Other recent works~\cite{hammerla2015pd, lipton2015learning} also leverage the power of deep learning approaches to model diseases and clinical time series data. These previous works have successfully showed the state-of-the-art performance of deep learning models for the healthcare domain but they have made limited attempts at the interpretability of the features learned by deep learning models, which prevents the clinician from understanding and applying these models.

As pointed out in the introduction, model interpretability is not only important but also \textit{necessary} in healthcare domain. Decision trees~\cite{quinlan1986induction} - due to their easy interpretability - have been quite successfully employed in the healthcare domain~\cite{bonner2001decision, yao2005r, fan2011hybrid} and clinicians have embraced it to make informed decisions. However, decision trees can easily overfit and they do not achieve good performance on datasets with missing values which is common in today's healthcare datasets. On the otherhand, deep learning models have achieved remarkable performance in healthcare as discussed in the previous paragraph, but their learned features are hardly interpretable. Here, we review some recent works on interpretability of deep learning features conducted in computer vision field.~\cite{erhan2009visualizing} studied the visualizations of the hierarchical representations learned by deep networks.~\cite{zeiler2014visualizing}  investigated not only the visualizations but also demonstrated the feature generalizablility in convolutional neural networks.~\cite{szegedy2013intriguing} argued that interpreting individual units can be misleading and postulated that the semantics encoded by hidden unit activations in one layer are preserved when projected onto random bases, instead of the next layer's bases. These works show that interpreting deep learning features is possible but the behavior of deep models may be more complex than previously believed. Therefore we believe there is a need to identify novel solutions which can provide interpretable models and achieve similar prediction performance as deep models.




Mimicking the performance of deep learning models using shallow models is a recent breakthrough in deep learning which has captured the attention of the machine learning community.~\cite{ba2014deep} showed empirically that shallow neural networks are capable of learning the same function as deep neural networks. They demonstrated this by first training a state-of-the-art deep model, and then training a shallow model to mimic the deep model. Motivated by the model compression idea from~\cite{bucilua2006model}, ~\cite{hinton2015distilling} proposed an efficient knowledge distillation approach to transfer (dark) knowledge from model ensembles into a single model.~\cite{korattikara2015bayesian} takes a Bayesian approach for distilling knowledge from a teacher neural network to train a student neural network. Excellent performance on real world tasks using mimic learning has been recently demonstrated in~\cite{li2014learning}.  All these previous works, motivate us to employ mimic learning strategy to learn an interpretable model from a well-trained deep neural network, which will be clearly discussed in the following section.

%
%
%
%

\section{Methods}
\label{sec:methods}

In this section, we will first describe the notations used in the paper, and then we describe the state-of-the-art deep learning models which we use as the original models and the Gradient Boosting Trees which we use as interpretable models for mimicking original models. Finally, we present the general pipeline of our interpretable mimic learning framework.

\subsection{Notations}
EHR data contains both static and temporal features. Without loss of generality, we assume each data sample has static records with $Q$ variables and temporal data of length $T$ and $P$ variables, as well as a binary label $y\in \{0,1\}$, where $ y $ usually represents a patient's health state. By flattening the time series and concatenating static variables, we get an input vector $\vct{X}\in {\R^D}$ for each sample, where $D=TP+Q$.

\subsection{Feedforward Network and Stacked Denosing Autoencoder}

A multilayer feedforward network~\cite{hornik1989multilayer} is a neural network with multiple nonlinear layers and possibly one prediction layer on the top.
The first layer takes $\vct{X}$ as the input, and the output of each layer is used as the input of the next layer. The transformation of each layer $l$ can be written as
\[
\vct{X}^{(l+1)}= f^{(l)}(\vct{X}^{(l)}) = s^{(l)}\left( \mat{W}^{(l)}\vct{X}^{(l)} + \vct{b}^{(l)} \right)
\]
where $\mat{W}^{(l)}$ and $\vct{b}^{(l)}$ are respectively the weight matrix and bias vector of layer $l$, and $s^{(l)}$ is a nonlinear activation function, which usually takes \textit{logistic sigmoid}, \textit{tanh}, or \textit{ReLU}~\cite{nair2010rectified} functions.
While we optimize the cross-entropy prediction loss and get the prediction output from topmost prediction layer, the activation of the hidden layers are also very useful as learned features.

The Stacked Autoencoder~\cite{bengio2007greedy} has a very similar structure as feedforward network, but its main objective is to minimize the squared reconstruction loss to the input instead of the cross-entropy prediction loss, by using encoder and decoder networks with tied weights. Assume the encoder has the same structure as feedforward network, then the $l$-th layer of the decoder takes the output $\vct{Z}^{(l+1)}$ from the next layer and transforms it by
\[
\vct{Z}^{(l)}= s^{(l)}\left( {\mat{W}^{(l)}}\T\vct{Z}^{(l+1)} + \vct{b_d}^{(l)} \right)
\]
where $\vct{Z}^{L+1} = \vct{X}^{(L+1)}$ is the output from the encoder, and finally $\vct{Z}^{(0)}$ can be treated as the reconstruction of the original input.
By adding noise to the input, i.e. hiding some input variables randomly, but still trying to recover the uncorrupted input, we obtain the Stacked Denoising Autoencoder~\cite{vincent2008extracting, vincent2010stacked} which is more robust to noise than the Stacked Autoencoder.
After training a stacked autoencoder, we add a logistic prediction layer on the encoder to solve the prediction task.

\subsection{Long Short-Term Memory}
\begin{figure}[htb]
\setlength\fboxsep{0pt}
\begin{center}
\subfigure[\label{fig:lstmblock}LSTM Block]{
\includegraphics[width=0.47\textwidth, trim={0 3.5in 5in 0}]{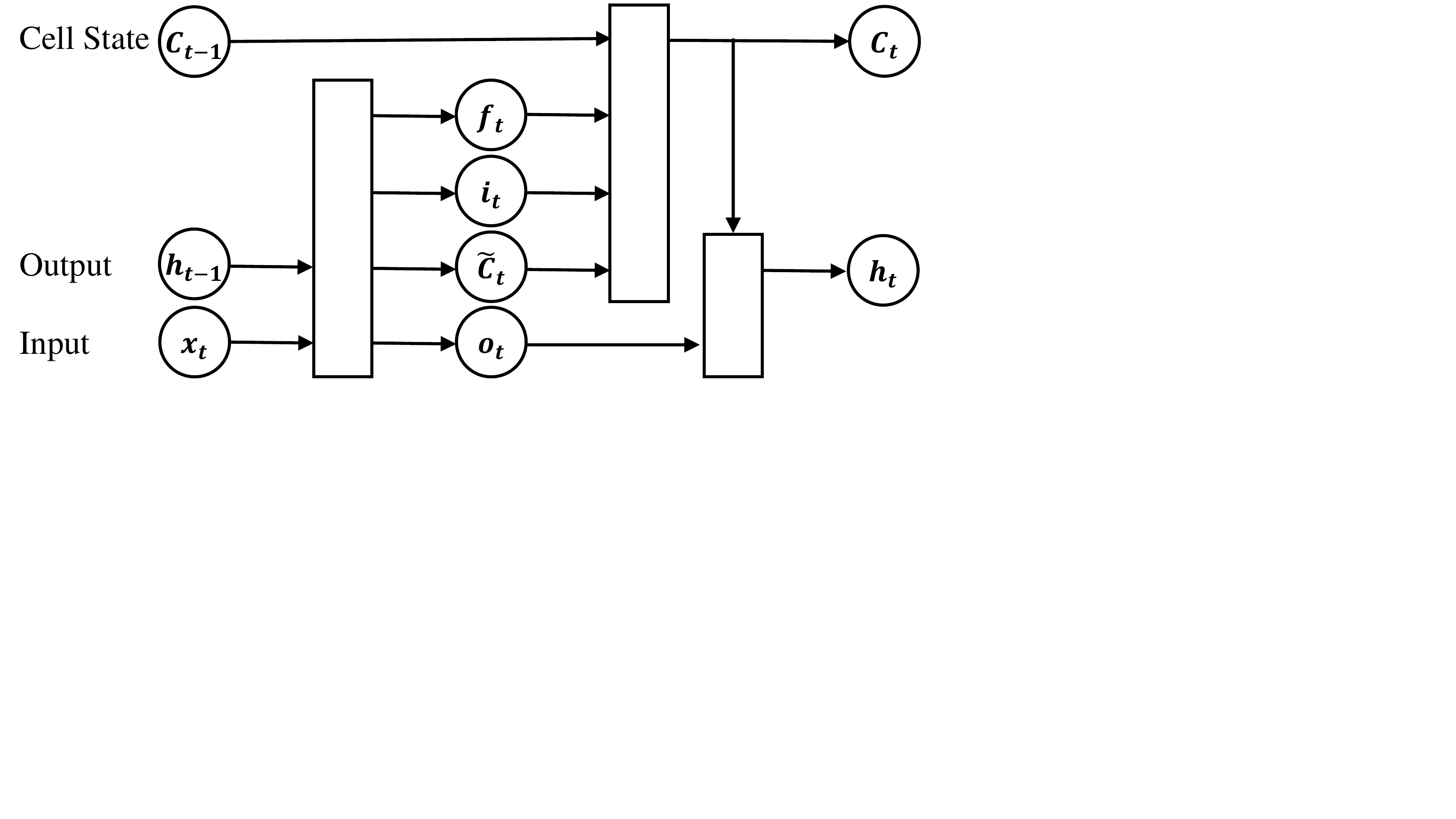}
}
\subfigure[\label{fig:lstmmodel}LSTM Prediction Model]{
\includegraphics[width=0.47\textwidth, trim={0 2.5in 6in 0}]{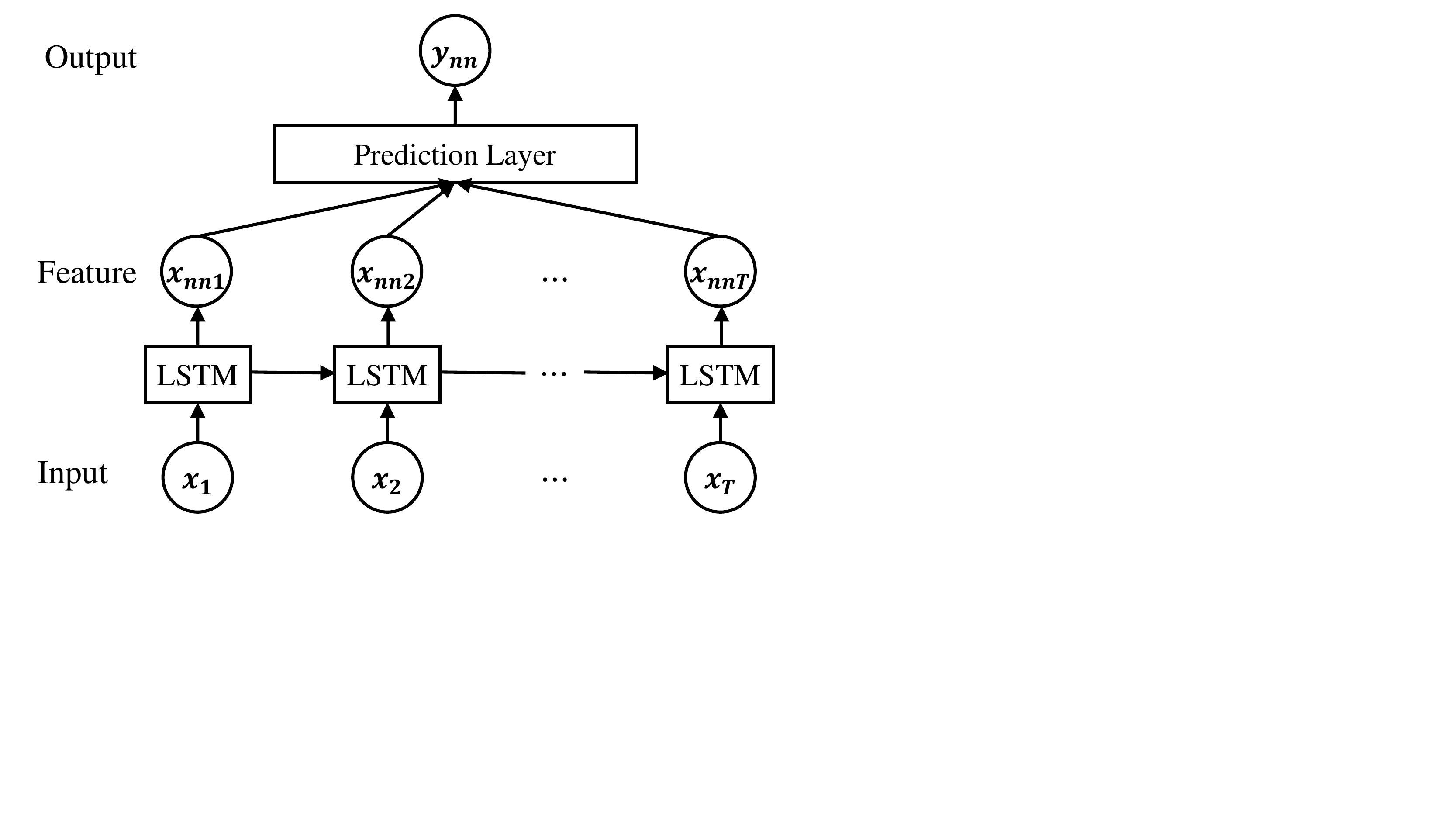}
}
\end{center}
\vspace{-0.2in}
\caption{A Sketch of Long Short-Term Memory Model}
\end{figure}

If we want to apply temporal model and only focus on the $P$ temporal variables, we can apply time series models on input $\mat{X}_{ts}={(x_1, x_2, \cdots, x_T)}\T\in\R^{T\times P}$, where $x_t \in \R^P$ is the variables at time $t$.
Long Short-Term Memory (LSTM)~\cite{hochreiter1997long} is a popular recurrent neural networks for sequential data and tasks.
It is used to avoid the vanishing gradient problem which is prevalent in other recurrent neural network architectures.
Figure~\ref{fig:lstmblock} shows the standard structure of an LSTM block with input, forget, and output gates, which we use in our method. In step $t$, one LSTM block takes the time series input $x_t$ at that time, cell state $C_{t-1}$ and output $h_{t-1}$ from previous time step, and calculates the cell state $C_t$ and output $h_t$ at this time step. We use the following steps to compute the output from each gate:
\begin{align*}
f_t &= \sigma\left( \mat{W}_{fh} h_{t-1} +  \mat{W}_{fx} x_{t} + b_f\right) &
i_t &= \sigma\left( \mat{W}_{ih} h_{t-1} +  \mat{W}_{ix} x_{t} + b_i\right)
\\
\tilde{C}_t &= tanh\left( \mat{W}_{Ch} h_{t-1} +  \mat{W}_{Cx} x_{t} + b_C \right) &
o_t &= \sigma\left( \mat{W}_{oh} h_{t-1} +  \mat{W}_{ox} x_{t} + b_o \right)
\end{align*}

And the outputs from this LSTM block is computed as:
\begin{align*}
C_t &= f_t * C_{t-1} + i_t * \tilde{C}_t &
h_t &= o_t * tanh(C_t)
\end{align*}
where $*$ refers to the element-wise multiplication of two vectors.

In our LSTM prediction model, we flatten the output from each block, which is denoted as $\mat{X}_{nn} = (x_{nn1}, {x}_{nn2},\cdots, {x}_{nnT}) = (h_1, h_2, \cdots, h_T)$, and we add another prediction layer on top of them. The model is shown in Figure~\ref{fig:lstmmodel}.

\subsection{Gradient Boosting Trees}
Gradient boosting~\cite{friedman2001greedy, friedman2002stochastic} is a method which takes an ensemble of weak learners, usually decision trees, to optimize a differentiable loss function by stages.
The basic idea is that the prediction function $F(x)$ can be approximated by a linear combination of several functions (under some assumptions), and these functions can be seeked using gradient descent approaches.
At each stage $m$, assume the current model is $F_m(x)$, then the Gradient Boosting method tries to find a weak model $h_m(x)$ to fit the gradient of the loss function with respect to $F(x)$ at $F_m(x)$. The coefficient $\gamma_m$ of the stage function is computed by the line search strategy to minimize the loss. The final model with $M$ stages can be written as
\[
F_{M}(x) = \sum_{i=1}^M \gamma_i h_i(x) + const
\]
In Gradient Boosting Trees, each weak learner is a simple classification or regression tree.
To keep gradient boosting from overfitting, a regularization method called shrinkage is usually employed, which multiplies a small learning rate $\nu$ to the stage function in each stage. In this situation, the term $\gamma_i h_i(x)$ in the update rule above is replaced by $\nu \gamma_i h_i(x)$.

\subsection{Interpretable Mimic Learning method}
\label{sec:methods-mimic}
\begin{figure}[htb]
\setlength\fboxsep{0pt}
\begin{center}
\includegraphics[height=1.25in, trim={0 4.5in 0 0}]{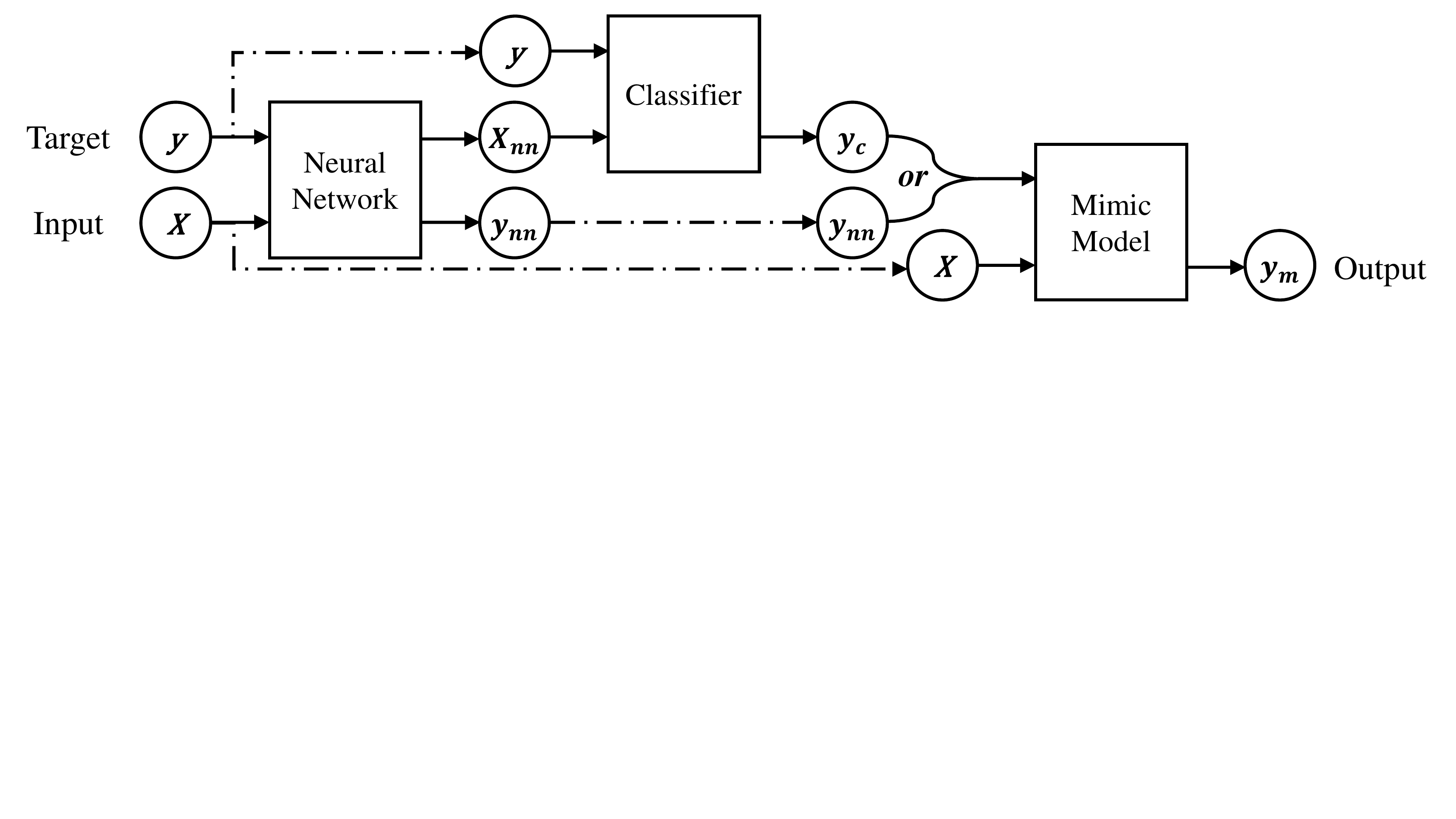}
\end{center}
\vspace{-0.2in}
\caption{\label{fig:mimicmethod}Training Pipeline for Mimic Method}
\end{figure}

In this section, we describe a simple but effective knowledge distillation framework - the \textit{Interpretable Mimic Learning} method also termed as the \textit{GBTmimic model}, which trains Gradient Boosting Trees to mimic the performance of deep network models. Our mimic method aims to recognize interpretable features while maintaining the state-of-the-art classification performance of the deep learning models.
To investigate, for different deep networks, whether using neural network extracted features rather than soft labels improve the mimicking performance we present two general pipelines for our GBTmimic model. 
The two pipelines of GBTmimic model are shown in Figure~\ref{fig:mimicmethod}.

In Pipeline 1, we utilize the learned features from deep networks and resort to another classifier such as Logistic Regression:

\begin{enumerate}
\item Given the input features $\vct{X}$ and target $y$, we train a deep neural network, either Stacked Denoising Autoencoder or Long Short-Term Memory, with several hidden layers and one prediction layer. We take the activations of the highest hidden layers as the extracted features $\vct{X}_{nn}$ from that deep network.
\item We then feed these new features into a standard classifier, e.g., Logistic Regression, to train on the same classification task, i.e. the target is still $y$, and take the soft prediction scores $y_c$ from the classifier.
\item Finally we train a mimic model, i.e., Gradient Boosting Regression Trees, given the raw input $\vct{X}$ and the soft targets $y_c$ to get the final output $y_m$ with minimum mean squared error.
\end{enumerate}

In Pipeline 2, we directly use the predicted soft-labels from deep networks.
\begin{enumerate}
\item The first step is similar to that in Pipeline 1, where we train a deep neural network with input features $\vct{X}$ and target $y$, but we take take the soft prediction scores $y_{nn}$ directly from the prediction layer of the neural network.
\item Instead of training an extra classifier with extracted features, we take the soft prediction scores $y_{nn}$ use it as the target in training the mimic model. In other words, we train Gradient Boosting Trees, which can output $y_m$ with minimum mean squared error to $y_{nn}$, given the raw input $\vct{X}$.
\end{enumerate}

After finishing the training procedure, we can directly apply the mimic model trained in the final step to the original classification task.
We compare these two different pipelines to investigate whether utilizing the features extracted from the neural networks (Pipeline 1) will provide more benefits than only taking the soft-labels from the neural networks (Pipeline 2), which is what existing mimic methods usually do.
These two pipelines will be evaluated and discussed with detailed experimental results in Section~\ref{sec:experiments}.

Our interpretable mimic learning model using GBT has several advantages over existing (mimic) methods.
First, gradient boosting trees is good at maintaining the performance of the original complex model such as deep networks by mimicing its predictions.
Second, it provides better interpretability than original model, from its decision rules and tree structures.
Furthermore, using soft targets from deep learning models avoids overfitting to the original data and provides good generalizations, which can not be achieved by standard decision tree methods.

%
%

\section{Experiments}
\label{sec:experiments}
We demonstrate the performance of our interpretable mimic learning framework on a real-world healthcare dataset and compare it to the several methods introduced in the previous section. Our experiments will help us to answer the following questions:

\begin{itemize}
	\item How does our interpretable mimic learning perform when compared with state-of-the-art deep learning and other machine learning methods?
	
	\item What are the interpretable features identified by our mimic learning framework?
	
	\item Do soft-labels from top layer of deep networks (Pipeline 2 in Section~\ref{sec:methods-mimic}) obtain better results than soft-labels of the same networks with Logistic Regression (Pipeline 1 in Section~\ref{sec:methods-mimic}) for prediction tasks?
	
	\item Do static features help in performance improvement in our mimic learning framework?
	
\end{itemize}

In the remainder of this section, we will describe the datasets, experimental design and discuss our empirical results and interpretations to answer the above questions.

\subsection{Dataset Descriptions}
We conducted a series of experiments on \textit{VENT dataset}~\cite{khemani2009effect}.
This dataset consists of data from 398 patients with acute hypoxemic respiratory failure in the intensive care unit at Children's Hospital Los Angeles (CHLA).
It contains a set of 27 static features, such as demographic information and admission diagnoses, and another set of 21 temporal features (recorded daily), including monitoring features and discretized scores made by experts, during the initial 4 days of mechanical ventilation. Two of the time series features start from 0 in time step 0, so when we flatten time series and concatenate all features together, we omit these two 0-valued features and obtain the input feature vector with length $27+21\times 4 - 2 = 109$.
The missing value rate of this dataset is 13.43\%, with some patients/variables having a missing rate of $ >30\%$ . We perform simple imputation for filling the missing values where we take the majority value for binary variables, and empirical mean for other variables.
Please see table \ref{tab:ventdataset} for a detailed summary of this dataset.
\begin{table}[htb]
\caption{VENT Dataset Details}
\label{tab:ventdataset}
\small
\begin{center}
\begin{tabular}{lll}
\bf Feature type & \bf Number of features & \bf Feature examples\\
\hline
\bf Static features & $27$ & PIM scores, Demographics, Admission diagnosis, etc.\\
\bf Temporal features & $21\times 4$ &  Injury markers, Ventilator settings, blood gas values, etc.\\
\end{tabular}
\end{center}
\vspace{-1.5em}
\end{table}

\subsection{Experimental Design}
We conduct two binary classification tasks on VENT dataset:
\begin{itemize}
	\item Mortality (MOR) task -- In this task we predict whether the patient dies within 60 days after admission or not. In the dataset, there are 80 patients with positive mortality label (patients who die).
	\item Ventilator Free Days (VFD) task -- In this task, we are interested in evaluating a surrogate outcome of morbidity and mortality (Ventilator free Days, of which lower value is bad), by identifying patients who survive and are on a ventilator for longer than 14 days. Since here lower VFD is bad, it is a bad outcome if the value $\le 14$, otherwise it is a good outcome.
In the dataset, there are 235 patients with positive VFD labels (patients who survive and stay long enough on ventilators).
	
\end{itemize}

We report Area Under ROC (AUC) as the evaluation metric to compare proposed and related methods.

\subsection{Methods and Implementation Details}
\label{sec:exp-methods}
We categorize the methods in our main experiments into three groups:
\begin{itemize}
\item Baseline machine learning algorithms which are popularly used in the healthcare domain: Linear Support Vector Machine (SVM), Logistic Regression (LR), Decision Tree (DT), and Gradient Boosting Trees (GBT).
\item Neural network-based methods (NN-based): Deep Feed-forward Neural Network (DNN), Stack Denoising Autoencoder (SDA), and Long Short-Term Memory (LSTM). Based on the two pipelines of our Interpretable Mimic Learning methods, we have two kinds of NN-based methods:
    \begin{itemize}
    \item Using the neural network models to directly make classification. We denote these methods as DNN, SDA, and LSTM. (Pipline 2 in Section~\ref{sec:methods-mimic})
    \item Taking the activations of the highest hidden layers of the networks as the output features, and feeding them into Logistic Regression to obtain final prediction. These methods are denoted by LR-* (LR-DNN, LR-SDA, LR-LSTM) in this section. (Pipline 1 in Section~\ref{sec:methods-mimic})
    \end{itemize}
\item Our Interpretable Mimic Learning methods: For each of the NN-based methods described above, we take its soft predictions and treat it as the training target of Gradient Boosting Trees. These methods are denoted by GBTmimic-* (E.g., GBTmimic-LSTM, GBTmimic-LR-SDA, etc). As a comparison, we also try Decision Tree (DTmimic-*) as the mimic method.
\end{itemize}

We train all the algorithms with 5 random trials of 5-fold cross validation.
Our DNN and SDA implementations have two hidden layers and one prediction layer. We set the size of each hidden layer twice as large as input size. We do 50 epochs of  stochastic gradient descent (SGD) with learning rate 0.001.
For LSTM, we only take the time series features to fit this model and do 50 epochs RMSprop~\cite{tieleman2012lecture} training with learning rate 0.001. We stack a prediction layer over the sequence output of LSTM. When we take the output features from the LSTM model, we take the flattened sequence output.
We implement all the deep networks in Theano~\cite{Bastien-Theano-2012} and Keras~\cite{cholletkeras} platforms on a desktop with 4-core CPU and 16GB RAM.
For Decision Trees, we expand the nodes as deep as possible until all leaves are pure.
For Gradient Boosting Trees, we use stage shrinking rate 0.1 and maximum number of boosting stages 100. We set the depth of each individual trees to be 3, i.e., the number of terminal nodes is no more than 8, which is fairly enough for boosting. We implement all decision tree based methods using the scikit-learn~\cite{scikit-learn} package.


\subsection{Quantitative Results}

Table~\ref{tab:pred-results} shows the prediction performance comparison of the models introduced in Section~\ref{sec:exp-methods}.
We observe that for both the classification tasks (MOR and VFD tasks), the deep models obtain better performance than standard machine learning baselines; and our interpretable mimic methods obtain similar or better performance than the deep models.
Our GBTmimic-LR-SDA and GBTmimic-LR-DNN obtains the best performance in MOR and VFD tasks, respectively.
We found that the predictions of DNN and SDA with Logistic Regression is better than just using the deep models, however this is not true for LSTM model.
One possible reason is that LSTM captures the temporal dependencies which help in prediction, while in other methods the time series are flattened during input and thus the temporal relations are not efficiently modeled. Similarly, the performance of our interpretable mimic learning methods always improve upon DNN and SDA, and are comparable to LSTM based methods.
\vspace{-1em}
\begin{table}[htb]
\centering
\caption{Classification Results.}
\small
AUC: Mean of Area Under ROC;\\
AUC(std): Standard Deviation of Area Under ROC.
\label{tab:pred-results}
\normalsize
\begin{tabular}{|c|l|c|c|c|c|}
\hline
\multicolumn{2}{|c|}{\multirow{3}{*}{\textbf{Method}}} & \multicolumn{4}{c|}{\textbf{Task}} \\ \cline{3-6}
\multicolumn{2}{|c|}{} & \multicolumn{2}{c|}{\textbf{MOR}} & \multicolumn{2}{c|}{\textbf{VFD}} \\ \cline{3-6}
\multicolumn{2}{|c|}{} & \textbf{AUC} & \textbf{AUC(std)} & \textbf{AUC} & \textbf{AUC(std)} \\ \hline
\multirow{4}{*}{\textbf{Baseline}} & \textbf{SVM} & 0.6431 & 0.059 & 0.7248 & 0.056 \\ \cline{2-6}
 & \textbf{LR} & 0.6888 & 0.068 & 0.7602 & 0.053 \\ \cline{2-6}
 & \textbf{DT} & 0.5965 & 0.081 & 0.6024 & 0.044 \\ \cline{2-6}
 & \textbf{GBT} & 0.7233 & 0.065 & 0.7630 & 0.051 \\ \hline
\multirow{6}{*}{\textbf{NN-based}} & \textbf{DNN} & 0.7288 & 0.084 & 0.7756 & 0.053 \\ \cline{2-6}
 & \textbf{SDA} & 0.7313 & 0.083 & 0.7211 & 0.051 \\ \cline{2-6}
 & \textbf{LSTM} & \textbf{0.7726} & 0.062 & 0.7720 & 0.061 \\ \cline{2-6}
 & \textbf{LR-DNN} & 0.7300 & 0.084 & 0.7759 & 0.052 \\ \cline{2-6}
 & \textbf{LR-SDA} & 0.7459 & 0.068 & \textbf{0.7818} & 0.051 \\ \cline{2-6}
 & \textbf{LR-LSTM} & 0.7658 & 0.063 & 0.7665 & 0.063 \\ \hline
\multirow{6}{*}{\textbf{Mimic}} & \textbf{GBTmimic-DNN} & 0.7574 & 0.064 & \textbf{0.7835} & 0.054 \\ \cline{2-6}
 & \textbf{GBTmimic-SDA} & 0.7382 & 0.084 & 0.7194 & 0.049 \\ \cline{2-6}
 & \textbf{GBTmimic-LSTM} & \textbf{0.7668} & 0.059 & 0.7357 & 0.054 \\ \cline{2-6}
 & \textbf{GBTmimic-LR-DNN} & \textbf{0.7673} & 0.070 & \textbf{0.7862} & 0.058 \\ \cline{2-6}
 & \textbf{GBTmimic-LR-SDA} & \textbf{0.7793} & 0.066 & \textbf{0.7818} & 0.049 \\ \cline{2-6}
 & \textbf{GBTmimic-LR-LSTM} & 0.7555 & 0.067 & 0.7524 & 0.060 \\ \hline
\end{tabular}
\end{table}
\vspace{-0.5em}

Based on the observations from the above prediction results in Table~\ref{tab:pred-results}, and by noticing that LSTM only takes temporal features, we investigated whether time series features themselves are sufficient for our prediction tasks (i.e. we do not consider static features in input vectors). In other words, it is useful to demostrate whether that the temporal models are more relevant than the just static models based on the initial settings.

We conducted two new sets of experiments, 1) with only temporal features as input, and 2) with only static features and the initial values of temporal features at day 0. We present the results of these experiments in Table~\ref{tab:xt-results} and Table~\ref{tab:xnt0-results}, respectively. 
From Table~\ref{tab:xt-results} we can notice that, for MOR task, the prediction differences between temporal and all features are quite small (i.e. AUC(diff)), while in VFD task, we find that adding static features is relatively more critical to the prediction performance.
The different behaviours on these two tasks also explain why LSTM performs better in MOR task than in VFD task. 
Note that we don't show the LSTM results in Table~\ref{tab:xt-results} since we have already used only temporal features for LSTM prediction task and the corresponding results is in Table~\ref{tab:pred-results}.
Results from Table~\ref{tab:xnt0-results} further verified the superiority of the temporal models over just the static model. For both MOR and VFD tasks, the performances of only static variables and initial values of temporal variables degraded significantly on all tested models, and are even worse than the models with only temporal features in Table~\ref{tab:xt-results}.

\begin{table}[htb]
\centering
\caption{Classification Results of Input with Only Temporal Features.}
\small
AUC: Mean of Area Under ROC; \\
AUC(diff): AUC of all features (Table~\ref{tab:pred-results}) - AUC of temporal features (Table~\ref{tab:xt-results});\\
AUC(std): Standard Deviation of Area Under ROC.
\label{tab:xt-results}
\begin{tabular}{|l|c|c|c|c|c|c|}
\hline
\multicolumn{1}{|c|}{\multirow{3}{*}{\textbf{Method}}} & \multicolumn{6}{c|}{\textbf{Task}} \\ \cline{2-7}
\multicolumn{1}{|c|}{} & \multicolumn{3}{c|}{\textbf{MOR}} & \multicolumn{3}{c|}{\textbf{VFD}} \\ \cline{2-7}
\multicolumn{1}{|c|}{} & \textbf{AUC} & \multicolumn{1}{l|}{\textbf{AUC(diff)}} & \textbf{AUC(std)} & \textbf{AUC} & \multicolumn{1}{l|}{\textbf{AUC(diff)}} & \textbf{AUC(std)} \\ \hline
\textbf{LR} & 0.7013 & -0.0125 & 0.064 & 0.7344 & 0.0258 & 0.069 \\ \hline
\textbf{GBT} & 0.7202 & 0.0031 & 0.068 & 0.7420 & 0.0210 & 0.048 \\ \hline
\textbf{DNN} & 0.7455 & -0.0167 & 0.071 & 0.7591 & 0.0165 & 0.068 \\ \hline
\textbf{SDA} & 0.7332 & -0.0019 & 0.082 & 0.7175 & 0.0036 & 0.053 \\ \hline
\textbf{LR-DNN} & 0.7453 & -0.0153 & 0.716 & 0.7590 & 0.0169 & 0.068 \\ \hline
\textbf{LR-SDA} & 0.7395 & 0.0064 & 0.741 & 0.7622 & 0.0196 & 0.053 \\ \hline
\textbf{GBTmimic-DNN} & 0.7632 & -0.0058 & 0.065 & 0.7579 & 0.0256 & 0.061 \\ \hline
\textbf{GBTmimic-SDA} & 0.7380 & 0.0002 & 0.083 & 0.7185 & 0.0009 & 0.056 \\ \hline
\textbf{GBTmimic-LSTM} & \textbf{0.7667} & 0.0001 & 0.058 & 0.7380 & -0.0023 & 0.050 \\ \hline
\textbf{GBTmimic-LR-DNN} & 0.7613 & 0.0060 & 0.055 & 0.7595 & 0.0267 & 0.066 \\ \hline
\textbf{GBTmimic-LR-SDA} & 0.7574 & 0.0219 & 0.060 & \textbf{0.7726} & 0.0092 & 0.056 \\ \hline
\textbf{GBTmimic-LR-LSTM} & 0.7565 & -0.0010 & 0.060 & 0.7263 & 0.0261 & 0.058 \\ \hline
\end{tabular}
\end{table}

\begin{table}[htb]
\centering
\caption{Classification Results of Input with Static and Initial Temporal (at Day 0) Features.}
\small
AUC: Mean of Area Under ROC; \\
AUC(diff): AUC of all features (Table~\ref{tab:pred-results}) - AUC of static and initial temporal features (Table~\ref{tab:xnt0-results});\\
AUC(diff2): AUC of temporal features (Table~\ref{tab:xt-results}) - AUC of static and initial temporal features (Table~\ref{tab:xnt0-results});\\
\label{tab:xnt0-results}
\begin{tabular}{|l|c|c|c|c|c|c|}
\hline
\multicolumn{1}{|c|}{\multirow{3}{*}{\textbf{Method}}} & \multicolumn{6}{c|}{\textbf{Task}} \\ \cline{2-7}
\multicolumn{1}{|c|}{} & \multicolumn{3}{c|}{\textbf{MOR}} & \multicolumn{3}{c|}{\textbf{VFD}} \\ \cline{2-7}
\multicolumn{1}{|c|}{} & \textbf{AUC} & \multicolumn{1}{l|}{\textbf{AUC(diff)}} & \textbf{AUC(diff2)} & \textbf{AUC} & \multicolumn{1}{l|}{\textbf{AUC(diff)}} & \textbf{AUC(diff2)} \\ \hline
\textbf{LR} & 0.6797 & 0.0091 & 0.0216 & 0.7158 & 0.0444 & 0.0186 \\ \hline
\textbf{GBT} & 0.6812 & 0.0421 & 0.0390 & 0.7081 & 0.0549 & 0.0339 \\ \hline
\textbf{DNN} & 0.7058 & 0.0230 & 0.0397 & 0.7214 & 0.0542 & 0.0377 \\ \hline
\textbf{SDA} & 0.6981 & 0.0332 & 0.0351 & 0.6919 & 0.0292 & 0.0256 \\ \hline
\textbf{LR-DNN} & 0.7067 & 0.0233 & 0.0386 & 0.7237 & 0.0522 & 0.0353 \\ \hline
\textbf{LR-SDA} & 0.6967 & 0.0492 & 0.0428 & 0.7317 & 0.0501 & 0.0305 \\ \hline
\textbf{GBTmimic-DNN} & \textbf{0.7378} & 0.0196 & 0.0254 & 0.721 & 0.0625 & 0.0369 \\ \hline
\textbf{GBTmimic-SDA} & 0.6979 & 0.0403 & 0.0401 & 0.6878 & 0.0316 & 0.0307 \\ \hline
\textbf{GBTmimic-LR-DNN} & 0.7305 & 0.0368 & 0.0308 & 0.7235 & 0.0627 & 0.0145 \\ \hline
\textbf{GBTmimic-LR-SDA} & 0.7296 & 0.0497 & 0.0278 & \textbf{0.7322} & 0.0496 & 0.0273 \\ \hline
\end{tabular}
\end{table}

\subsection{Interpretations}

One advantage of decision tree methods is their interpretable feature selection and decision rules. In this section, we first interpret the trees learned by our mimic framework, and then we compare and contrast trees from our GBTmimic with trees obtained using original GBT.

Table~\ref{tab:imp-results} shows the top useful features, found by GBT and our GBTmimic models, in terms of the importance scores among all cross validations. We find that some important features are shared with several methods in these two tasks, e.g., MAP (Mean Airway Pressure) at day 1, $\delta$PF (Change of PaO2/FIO2 Ratio) at day 1, etc. Another interesting finding is that almost all the top features are temporal features, while among all static features, the PIM2 (Pediatric Index of Mortality 2) and PRISM (Pediatric Risk of Mortality) scores, which are developed and widely used by the doctors and medical experts, are the most useful variables.

\begin{table}[htb]
\centering
\caption{Top Features and Corresponding Importance Scores.}
Bold lines refer to the methods with the best classification results in that task.
\label{tab:imp-results}
\subtable[MOR Task]{
\centering
\small
\begin{tabular}{|l|l|l|l|l|}
\hline
\multicolumn{1}{|c|}{\textbf{Model}} & \multicolumn{4}{c|}{\textbf{Features (Importance Scores)}} \\ \hline
\textbf{GBT} & MAP-D1(0.052) & PaO2-D2(0.052) & FiO2-D3(0.037) & PH-D3(0.027) \\ \hline
\textbf{GBT-DNN} & MAP-D1(0.031) & $\delta$PF-D1(0.031) & PH-D1(0.029) & PIM2S(0.027) \\ \hline
\textbf{GBT-SDA} & OI-D1(0.036) & MAP-D1(0.032) & OI-D0(0.028) & LIS-D0(0.028) \\ \hline
\textbf{GBT-LSTM} & $\delta$PF-D1(0.058) & MAP-D1(0.053) & BE-D0(0.043) & PH-D1(0.042) \\ \hline
\textbf{GBT-LR-DNN} & $\delta$PF-D1(0.032) & PRISM12ROM(0.031) & PIM2S(0.031) & Unplanned(0.030) \\ \hline
\textbf{GBT-LR-SDA} & \textbf{PF-D0(0.036)} & \textbf{$\delta$PF-D1(0.036)} & \textbf{BE-D0(0.032)} & \textbf{MAP-D1(0.031)} \\ \hline
\textbf{GBT-LR-LSTM} & $\delta$PF-D1(0.066) & PH-D1(0.044) & MAP-D1(0.044) & PH-D3(0.041) \\ \hline
\end{tabular}
}
\subtable[VFD Task]{
\centering
\small
\begin{tabular}{|l|l|l|l|l|}
\hline
\multicolumn{1}{|c|}{\textbf{Model}} & \multicolumn{4}{c|}{\textbf{Features (Importance Scores)}} \\ \hline
\textbf{GBT} & MAP-D1(0.035) & MAP-D3(0.033) & PRISM12ROM(0.030) & VT-D1(0.029) \\ \hline
\textbf{GBT-DNN} & MAP-D1(0.042) & PaO2-D0(0.033) & PRISM12ROM(0.032) & PIM2S(0.030) \\ \hline
\textbf{GBT-SDA} & LIS-D0(0.049) & LIS-D1(0.039) & OI-D1(0.036) & PF-D3(0.032) \\ \hline
\textbf{GBT-LSTM} & $\delta$PF-D1(0.054) & MAP-D1(0.049) & PH-D1(0.046) & BE-D0(0.040) \\ \hline
\textbf{GBT-LR-DNN} & \textbf{PaO2-D0(0.047)} & \textbf{PIM2S(0.038)} & \textbf{MAP-D1(0.038)} & \textbf{VE-D0(0.034)} \\ \hline
\textbf{GBT-LR-SDA} & PaO2-D0(0.038) & VE-D0(0.034) & PH-D3(0.030) & MAP-D1(0.030) \\ \hline
\textbf{GBT-LR-LSTM} & PH-D3(0.062) & PaO2-D0(0.055) & $\delta$PF-D1(0.043) & MAP-D1(0.037) \\ \hline
\end{tabular}
}
\end{table}

\begin{figure}[htb]
\setlength\fboxsep{0pt}
\centering
\subfigure[GBT]{
\includegraphics[height=1.4in, trim={0.4in 0.4in 0.4in 0.4in},clip]{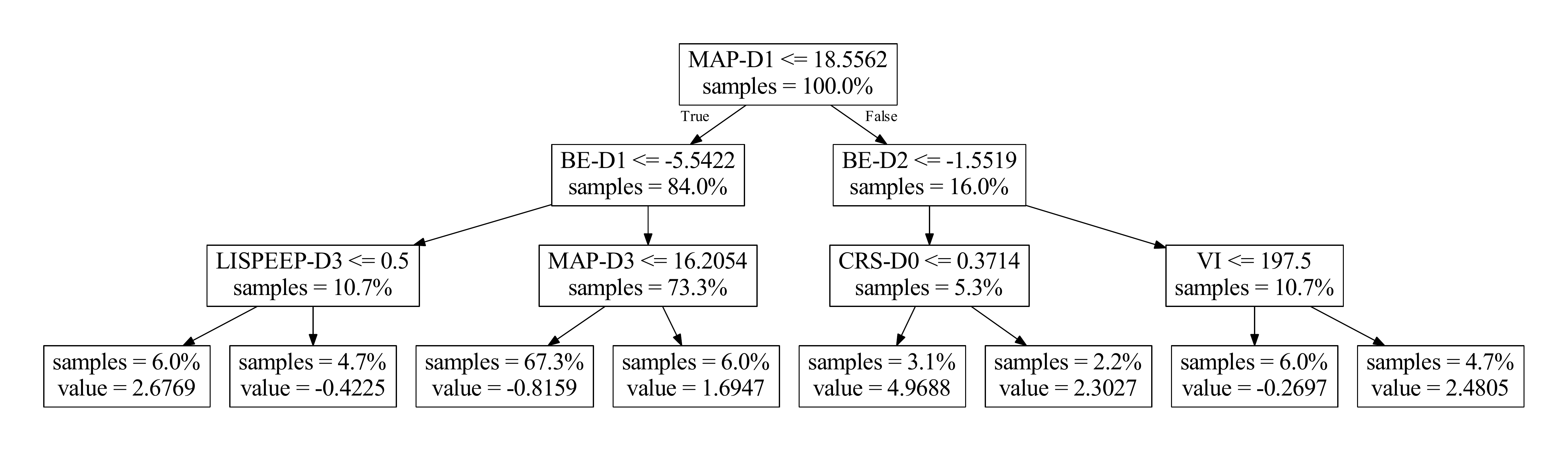}
}
\subfigure[GBTmimic-LR-SDA]{
\includegraphics[height=1.4in, trim={0.4in 0.4in 0.4in 0.4in},clip]{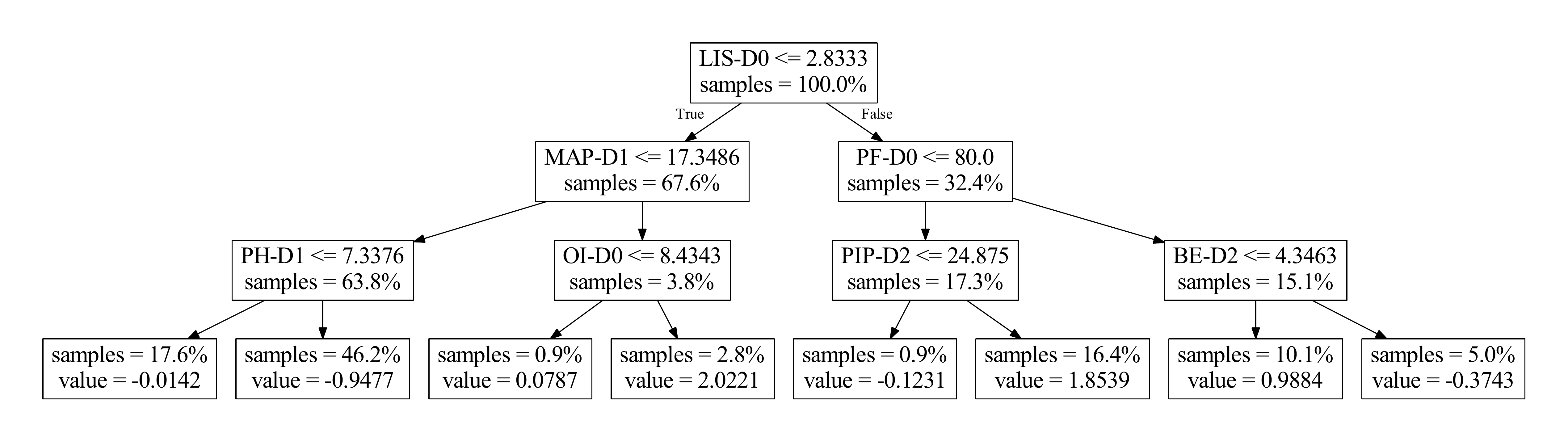}
}
\subfigure[GBTmimic-LSTM]{
\includegraphics[height=1.4in, trim={0.4in 0.4in 0.4in 0.4in},clip]{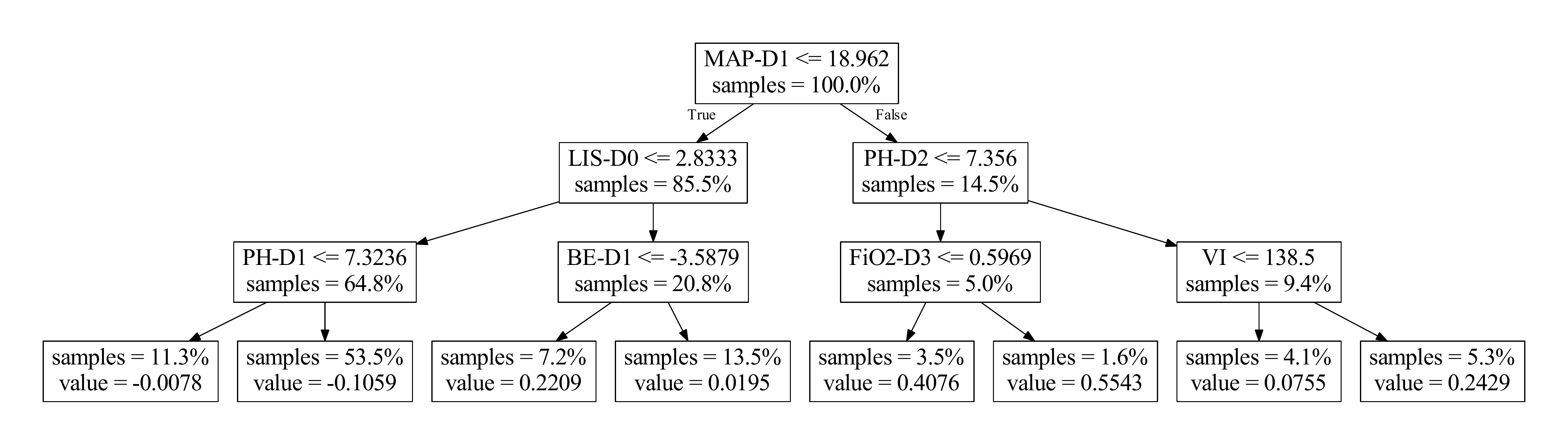}
}
\vspace{-0.2in}
\caption{\label{fig:tree}Important Decision Trees on MOR Task}
\small Value of a leaf node: The prediction score of a sample from the corresponding decision rules (path).
\end{figure}

We can compare and interpret the trees obtained by our Interpretable Mimic learning with the original GBT trees. Figure~\ref{fig:tree} shows the examples of the most important tree built by the original GBT and our interpretable mimic learning methods on the same cross validation fold for the MOR prediction task.
As we can see, they share some common features and similar rules. These selected features and rules can be evaluated and explained by healthcare experts which will help them to understand these models better and to make better decisions on patients.

Gradient Boosting Trees method has several internal tree estimators. Because of this, it can be more complex and harder to interpret than a single decision tree since a decision tree can be used to find one decision path for a single data sample.
So, we compare our GBTmimic-* with DTmimic-* which is obtained by mimicking a single decision tree. From Table~\ref{tab:dtmimic-results} we notice that DTmimic-* methods perform poorly compared to GBTmimic-* methods, which is not satisfying even if it (a single decision tree) can be better interpreted and visualized.

\begin{table}[htb]
	\centering
	\caption{Comparison of Mimic Methods with Decision Tree and Gradient Boosting Trees.}
	\small
	AUC: Mean of Area Under ROC; \\
	AUC(diff): AUC of GBTmimic (Table~\ref{tab:pred-results}) - AUC of DTmimic (Table~\ref{tab:dtmimic-results});\\
	AUC(std): Standard Deviation of Area Under ROC.
	\label{tab:dtmimic-results}
	\begin{tabular}{|l|c|c|c|c|c|c|}
		\hline
		\multicolumn{1}{|c|}{\multirow{3}{*}{\textbf{Method}}} & \multicolumn{6}{c|}{\textbf{Task}} \\ \cline{2-7}
		& \multicolumn{3}{c|}{\textbf{MOR}} & \multicolumn{3}{c|}{\textbf{VFD}} \\ \cline{2-7}
		& \textbf{AUC} & \multicolumn{1}{l|}{\textbf{AUC(diff)}} & \textbf{AUC(std)} & \textbf{AUC} & \multicolumn{1}{l|}{\textbf{AUC(diff)}} & \textbf{AUC(std)} \\ \hline
		\textbf{DTmimic-DNN} & 0.6683 & 0.0891 & 0.079 & 0.6769 & 0.1152 & 0.062 \\ \hline
		\textbf{DTmimic-SDA} & 0.7138 & 0.0244 & 0.087 & 0.7058 & 0.0056 & 0.053 \\ \hline
		\textbf{DTmimic-LSTM} & 0.7117 & 0.0551 & 0.076 & 0.6898 & 0.0240 & 0.054 \\ \hline
		\textbf{DTmimic-LR-DNN} & 0.6931 & 0.0742 & 0.068 & 0.6992 & 0.0931 & 0.048 \\ \hline
		\textbf{DTmimic-LR-SDA} & 0.6994 & 0.0799 & 0.072 & 0.6933 & 0.0824 & 0.063 \\ \hline
		\textbf{DTmimic-LR-LSTM} & 0.6976 & 0.0579 & 0.075 & 0.7098 & 0.0548 & 0.057 \\ \hline
	\end{tabular}
\end{table}

\section{Discussions}
\label{sec:discussions}
In this paper, we proposed a novel knowledge distillation approach from deep networks via Gradient Boosting Trees, which can be used to learn interpretable features and prediction rules.
Our preliminary experimental results show similar or even better performance from our mimic methods on a real world dataset, and demonstrate a very promising direction for future machine learning research in healthcare domain.

For future work, we aim to extract more useful information like decision rules or tree node features from our mimic methods for better diagnosis interpretability. We will also apply our proposed approaches on a larger healthcare dataset, such as MIMIC-II~\cite{saeed2011multiparameter} which is derived from multiple clinical sources, to further verify our methods. We also plan to extend our mimic methods to other  state-of-the-art machine learning models, such as structured deep network models, to explore their application abilities in difficult practical applications and help domain experts have a better understanding of these models.

\small
\bibliography{references}
\bibliographystyle{abbrv}

\end{document}